\documentclass[lettersize,journal]{IEEEtran}
\usepackage{amsmath,amsfonts}
\usepackage{algorithmic}
\usepackage{algorithm}
\usepackage{array}
\usepackage[caption=false,font=normalsize,labelfont=sf,textfont=sf]{subfig}
\usepackage{textcomp}
\usepackage{stfloats}
\usepackage{url}
\usepackage{verbatim}
\usepackage{graphicx}
\usepackage{cite}
\usepackage{multirow}
\usepackage[colorlinks,
            linkcolor=blue,
            anchorcolor=blue,
            citecolor=blue]{hyperref}
\usepackage{float}
\begin{document}

\title{NCAGC: A Neighborhood Contrast Framework for Attributed Graph Clustering}

\author{Tong Wang, Guanyu Yang,~\IEEEmembership{Senior Member,~IEEE,} Qijia He, Zhenquan Zhang, and Junhua Wu
\thanks{Tong Wang, Guanyu Yang and Zhenquan Zhang are with the Laboratory of Image Science and Technology and the Key Laboratory of Computer Network and Information Integration, Ministry of Education, Southeast University, Nanjing 210096, China (e-mail: tongwangnj@qq.com, yang.list@seu.edu.cn, zhangzhenq@foxmail.com).}
\thanks{Qijia He is with the CAS Key Laboratory of FAST, National Astronomical Observatory, Chinese Academy of Sciences, Beijing 100101, China (e-mail: qjhe@nao.cas.cn).}
\thanks{Junhua Wu is  with the Department of School of Computer Science and Technology, Nanjing Tech University, Jiangsu 211816, China (e-mail: wujh@njtech.edu.cn).}
}



\maketitle

\begin{abstract}
Attributed graph clustering is one of the most fundamental tasks among graph learning field, the goal of which is to group nodes with similar representations into the same cluster without human annotations. Recent studies based on graph contrastive learning method have achieved remarkable results when exploit graph-structured data. However, most existing methods 1) do not directly address the clustering task, since the representation learning and clustering process are separated; 2) depend too much on data augmentation, which greatly limits the capability of contrastive learning; 3) ignore the contrastive message for clustering tasks, which adversely degenerate the clustering results. In this paper, we propose a \underline{N}eighborhood \underline{C}ontrast Framework for \underline{A}ttributed \underline{G}raph \underline{C}lustering, namely NCAGC, seeking for conquering the aforementioned limitations. Specifically, by leveraging the Neighborhood Contrast Module, the representation of neighbor nodes will be 'push closer' and become clustering-oriented with the neighborhood contrast loss. Moreover, a Contrastive Self-Expression Module is built by minimizing the node representation before and after the self-expression layer to constraint the learning of self-expression matrix. All the modules of NCAGC are optimized in a unified framework, so the learned node representation contains clustering-oriented messages. Extensive experimental results on four attributed graph datasets demonstrate the promising performance of NCAGC compared with 16 state-of-the-art clustering methods. The code is available at \url{https://github.com/wangtong627/NCAGC}.
\end{abstract}

\begin{IEEEkeywords}
Attributed graph clustering, graph representation learning, contrastive learning.
\end{IEEEkeywords}

\section{Introduction}
\IEEEPARstart{E}ntering the information era, multimedia data gradually plays a critical role in people's daily life. The attributed graph, one of the most common style of multimedia data, is a graph with node attributes and widely exists in our real world, e.g., social network\cite{8579104, 8613906}, citation network\cite{kipf2016semi} and recommendation system\cite{ying2018graph, 9535249}. Attributed graph clustering\cite{cai2018comprehensive, 9181470, DBLP:conf/ijcai/ChengWTXG20, DBLP:conf/aaai/GaoXWXZ20} is a fundamental and active task in graph data mining, which aims at grouping the given nodes into several disjoint clusters in an unsupervised manner. However, considering the high-dimensional node attributes and complex graph structures, it is of great challenge for the attributed graph clustering methods to effectively exploit graph data. Traditional clustering methods such as K-Means\cite{hartigan1979algorithm} only care about node attributes, while Spectral \cite{ng2001spectral} method only focus on the graph structure, degrading the clustering performance.

To tackle this challenge, Graph Neural Network (GNN)\cite{kipf2016semi} has attracted considerable attention, which aims to extract node representation into low-dimension and simultaneously keep the original graph structures and node attributes as much as possible. Due to the powerful graph representation learning capability of GNN, various GNN-based unsupervised learning methods \cite{kipf2016variational, pan2018adversarially, 2020Learning, salehi2019graph, wang2019attributed, bo2020structural, wang2021multi, 9472979, 9508843, DBLP:journals/nn/XiaWYGHG22, DBLP:journals/nn/XuXGHG21} have been proposed. Among them, GAE and VGAE\cite{kipf2016variational} make an early attempt by employing a GCN encoder and a reconstruction decoder to exploit node representation for clustering. For better performance, ARGAE and ARVGAE\cite{pan2018adversarially} \cite{2020Learning} are proposed by adopting an adversarial training strategy based on \cite{kipf2016variational}. Besides, GATE\cite{salehi2019graph} utilizes an attention mechanism to reconstruct the node relation in the representation learning process. Although the aforementioned methods show superiority in exploiting graph-structured data, however, most of these methods are not designed for clustering tasks. To be precise, they just adopt GNN as a feature extractor and then use the traditional clustering method on the extracted node embedding. As the representation learning and clustering process of these methods are executed separated, the performance is also limited.

To build a clustering-oriented framework, many end-to-end graph clustering methods, which integrate representation learning and clustering task into a unified framework have been proposed. DAEGC\cite{wang2019attributed} obtains the clustering oriented node representation by optimizing the clustering distribution and target distribution. After that, SDCN\cite{bo2020structural} integrates an auto-encoder and a graph convolutional network into a unified framework and obtains the node embedding with graph structure information for clustering. MSGA\cite{wang2021multi} develops a multi-scale self-expression layer for graph clustering and uses the pseudo labels to guide the representation learning. Though these methods show commendable clustering performance, they cannot effectively exploit the heterogeneous node embedding, leading to inferior performance.

Recent studies have the most focus on contrastive learning\cite{9732218, he2020momentum, chen2020simple, van2020scan, 9772930, 9712249}, which achieves great success in many fields, e.g., computer vision and natural language processing, because of its powerful unsupervised learning ability. Inspired by this, many graph contrastive learning based methods \cite{9758652, zhu2020deep, zhu2021graph, DBLP:conf/nips/YouCSCWS20} attract much attention. Among them, GRACE\cite{zhu2020deep} obtains node representations by performing an instance-level graph contrastive learning that maximizes the consistency of the representation of the same node in two augmented views. Moreover, GCA\cite{zhu2021graph} proposes an adaptive graph augmentation method with node centrality measures based on \cite{zhu2020deep} and enhances the graph information for contrasting. Unfortunately, these graph contrastive learning based methods still face the following issues when they are applied to the attributed graph clustering task.

\begin{enumerate}
\item{They do not have a clustering-oriented loss function, since the aforementioned methods are just treated as a feature extractor. Because of this, the representation learning and clustering task are separated, resulting in the suboptimal clustering performance.}
\item{They rely too much on data augmentation, which is of crucial importance to the above graph contrastive method. However, due to the complexity of graph-structured data, it is uncertain whether the pre-defined graph data augmentation method is suitable for the corresponding dataset and down-stream tasks.}
\item{They merely utilize contrastive learning in representation learning, ignoring the contrastive message for down-stream tasks.}
\end{enumerate}

To conquer the aforementioned limitations, we propose a generic {\bf N}eighborhood {\bf C}ontrast Framework for {\bf A}ttributed {\bf G}raph {\bf C}lustering (NCAGC), which is characterized by two graph contrastive learning based modules. To be exact, we first employ a Symmetric Feature Extraction Module for extracting the node representation in an unsupervised manner. Besides, considering the nodes grouped in the same cluster should be more similar, we propose a Neighborhood Contrast Module to maximize the representation of neighbor nodes which are obtained by the KNN method. Considering the neighbor nodes should have similar representation, all the neighbor nodes are defined as positive samples leaving others to be negative, in this way, the learned node representation could be more suitable for the clustering task. Moreover, we propose a Contrastive Self-Expression Module to make good use of the node representation and construct the self-expression coefficient matrix for spectral clustering. By minimizing the node representation before and after the reconstruction of the self-expression layer in an instance-level contrast manner, we can learn a more discriminative linear combination coefficient, i.e., the self-expression coefficient matrix, which directly improves the clustering performance. By iteratively training and optimizing all the modules in a unified framework, the node representation learning and clustering tasks can benefit from each other. Extensive experiments over four attributed graph datasets demonstrate the superiority of our proposed NCAGC in terms of 16 state-of-the-art clustering methods. The main highlights of this paper is summarized as follows.

\begin{enumerate}
\item{A dual contrastive learning based attributed graph clustering method termed NCAGC is proposed, which employs a Neighborhood Contrast Module to enhance the learning of node representation making it caters to downstream clustering tasks.}
\item{To the best of our knowledge, NCAGC could be the first method that utilizes a contrastive learning strategy for learning a self-expression matrix in the Contrastive Self-Expression Module. Different from other self-expression based subspace clustering models, NCAGC takes account both the corresponding nodes, i.e., positive pairs, and other nodes, i.e., negative pairs, before and after the self-expression layer, which helps to learn a more discriminative self-expression matrix for clustering.}
\item{Different from other contrastive learning methods, NCAGC selects positive/negative pairs in the original view without data augmentation. Considering the complexity of graph data, it is still uncertain what kind of pre-defined graph data augmentation can satisfy the given data, since inappropriate augmentation may degrade the performance.}
\item{Benefiting from two contrastive learning based modules, NCAGC can learn high-quality representations and a discriminative self-expression matrix for clustering. Extensive experiments show that NCAGC outperforms 16 SOTA clustering methods according to three metrics.}
\end{enumerate}

\section{Related work}
In this section, we briefly introduce some recent developments in two related topics, namely, attributed graph clustering and graph contrastive learning.

\subsection{Attributed Graph Clustering}
Attributed graph clustering which aims to divide node samples into different disjoint clusters, is an essential task to explore the information embedded in the network-structured data. Although promising results have been achieved, traditional clustering methods such as K-Means and Spectral method give discouraging results on complex graph-structured data due to the ability to extract node representation. Benefit from the ability to capture complex and non-linear node representations, Graph Neural Network (GNN) based clustering methods have been widely addressed in recent years. Typically, these methods apply the traditional clustering method to the node embedding extracted by GNN. In particular, GAE adopts a two-layer GCN encoder and reconstructs the graph structure with an inner-product decoder. After that, many methods like ARGAE, DAEGC, and MGAE improve the clustering performance by introducing different strategies, such as adversarial training, attention mechanism, marginalized method, etc. Although the aforementioned methods achieve considerable performance, the lack of utilizing self-supervised information still limits the result. More recently, MSGA is proposed by integrating multiscale and self-supervised information in the representation process. Although MSGA proves that using self-supervised learning can effectively enhance the quality of node representation, in this manner, updating the node representation with the pseudo labels may still affect the convergence of the model. To this end, we introduce the contrastive learning method, a promising paradigm of self-supervised learning, for facilitating node representation learning and attributed clustering tasks.

\subsection{Graph Contrastive Learning}
Benefiting from its powerful unsupervised learning ability, contrastive learning has made considerable achievement\cite{he2020momentum, chen2020simple, van2020scan} in representation learning, computer vision, and natural language processing. The goal of contrastive learning is to map the raw data into a contrastive representation space where the similarity of positive pairs should be maximized and that of negative pairs should be minimized\cite{hadsell2006dimensionality}. As previously discussed, various works introduce contrastive learning methods into graph representation learning tasks and these works are called graph representation learning methods. Most of the existing graph representation learning networks treat the contrastive instance through the graph augmentation\cite{zhu2020deep, zhu2021graph, you2020graph, xia2021self, DBLP:conf/nips/YouCSCWS20}. To be specific, the positive pair is the same instance from two augmented views, while others are defined to be negative. Among them, GRACE proposes a graph contrast representation learning network with graph augmentation approaches and the local node-level instance contrastive loss function. To get better contrast quality, GCA improves the graph augmentation approach with an adaptive method, which highlights the important connective structures and adds more noise to the unimportant nodes.
Different from the above graph contrastive learning methods, NCAGC makes the contribution in the following aspects. First, the existing works depend on the data augmentation to contrast different pairs, however, it is not clear whether the pre-defined data augmentation is suitable for the corresponding tasks, whereas our method adopts another way to contrast different pairs without data augmentation. Second, the existing works perform contrastive learning methods simply on a single node or graph, whereas our method applies it to the selected neighborhood in the representation learning process. Last, the existing methods aim to learn the node representation while our method is designed for attributed graph clustering with contrastive self-expression loss.

\begin{figure*}[t]
\centering
\includegraphics[width=1.0\textwidth]{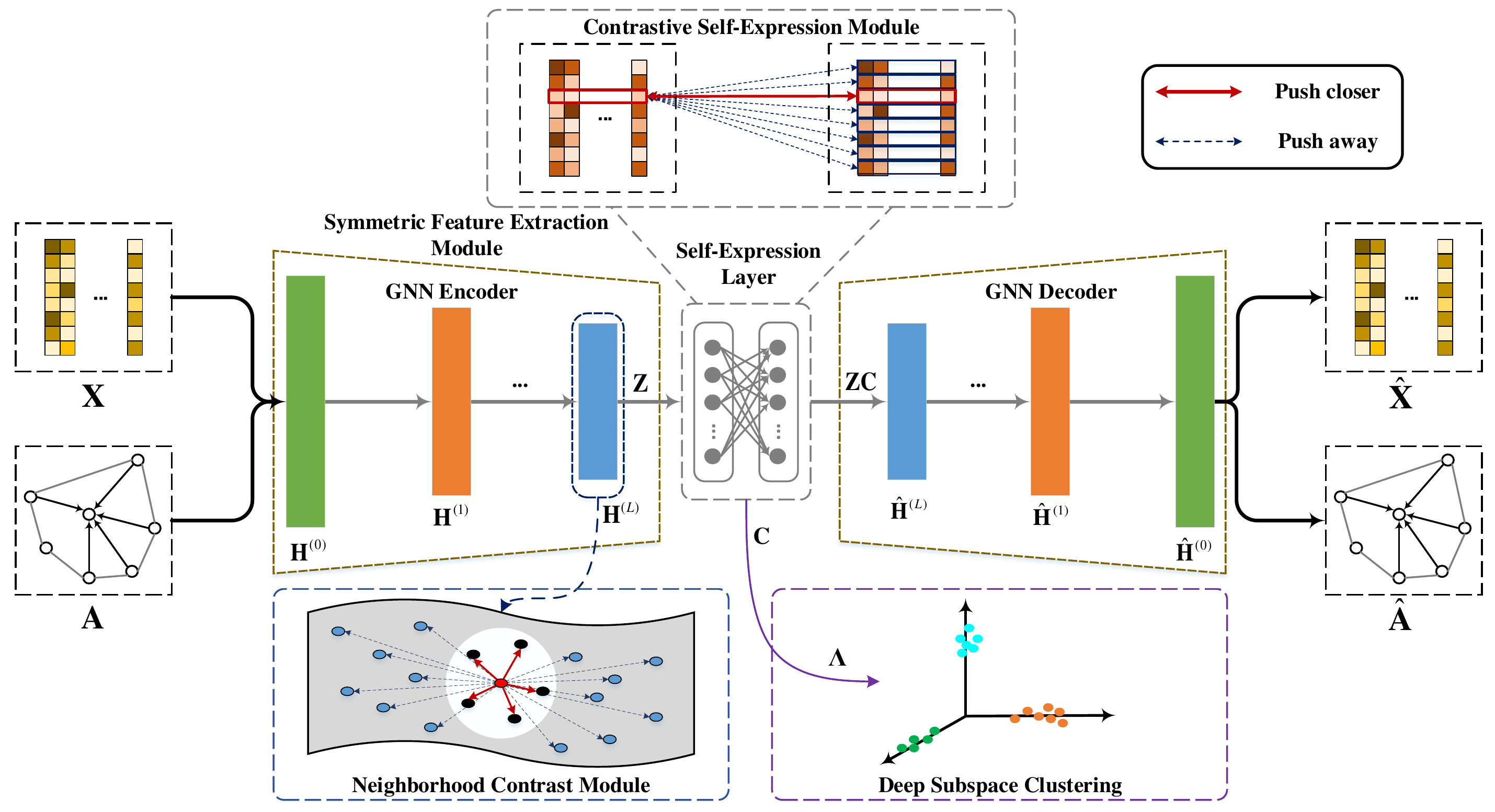}%
\caption{General Framework of our proposed NCAGC, which consists three modules: 1) Symmetric Feature Extraction Module. 2) Neighborhood Contrast Module. 3) Contrastive Self-Expression Module. $\mathbf{X}$ is the node attribute matrix, $\mathbf{A}$ is the adjacency matrix, $\mathbf{H}^{(i)}$ is the latent node representation in $i$-th layer of the GNN feature extractor, $\mathbf{C}$ is the self-expression matrix and $\mathbf{\Lambda}$ is the affinity matrix.}
\label{fig1}
\end{figure*}

\section{Methodology}
In this section, we formally introduce our proposed NCAGC, where the overall framework is shown in Fig. \ref{fig1}. Specifically, we employ two modules using contrastive learning, i.e., the Neighborhood Contrast Module and the Contrastive Self-Expression Module. We will describe them in detail in the following.

\subsection{Notations}
Given an undirected attribute graph $\mathcal{G}=(\mathbf{V}, \mathbf{E}, \mathbf{X})$, where $\mathbf{V}=\{v_1, v_2, ..., v_N\}$ and $\mathbf{E}$ are the node set and edge set, $\mathbf{X}=\{\mathbf{x}_1, \mathbf{x}_2, ..., \mathbf{x}_N\}$ is the node attribute matrix with $d$ features and $N$ is the number of nodes and $\mathbf{x}_i\in \mathbb{R}^d$ corresponding to the $i$-th column of matrix $\mathbf{X}\in \mathbb{R}^{d \times N}$. $\mathbf{A}\in \mathbb{R}^{N \times N}$ is the adjacency matrix and $\mathbf{A}_{ij}=1$ iff $(v_i,v_j) \in \mathbf{E}$ , i.e. there is an edge between node $v_i$ and $v_j$, otherwise $\mathbf{A}_{ij}=0$. $\mathbf{H}=[\mathbf{h}_1, \mathbf{h}_2, ..., \mathbf{h}_N]$ is the node representation matrix where $\mathbf{h}_i \in \mathbb{R}^ {d^{(j)}}$ corresponding to the $i$-th column of matrix $\mathbf{H}\in \mathbb{R}^{d^{(j)} \times N}$ and $d^{(j)}$ is the dimension of the node representation.

The goal of the attributed graph clustering is to divide the $N$ unlabeled nodes into $k$ disjoint clusters, providing the nodes in the same cluster have high similarity according to their attribute and adjacency information.

\subsection{Overall Framework}
As shown in Fig. \ref{fig1}, our proposed NCAGC contains three joint optimized components, namely Symmetric Feature Extraction Module, Neighborhood Contrast Module, and Contrastive Self-Expression Module.

\begin{itemize}
\item{{\bf{Symmetric Feature Extraction Module}} employs a symmetric GNN to encode the latent node representation from the node attribute and graph structure simultaneously and applies it to downstream clustering tasks.}
\item{{\bf{Neighborhood Contrast Module}} utilizes a contrastive learning method to improve the quality of extracted node representation by maximizing the similarities of top K nearest neighbor nodes, i.e., positive pairs, and minimizing the similarities of other nodes, i.e., negative pairs.}
\item{{\bf{Contrastive Self-Expression Module}} is leveraged to help learn a more discriminative self-expression coefficient matrix by contrasting the node representation before and after the reconstruction of the self-expression layer.}
\end{itemize}

\subsection{Symmetric Feature Extraction Module}
As present in Fig. \ref{fig1}, the Symmetric Feature Extraction Module includes a GNN encoder and a symmetric decoder to map the raw node attribute and graph structure into a new low-dimensional space for the attribute graph clustering.

Given the node attribute matrix $\mathbf{X}\in \mathbb{R}^{d \times N}$ and the graph structure matrix $\mathbf{A}\in \mathbb{R}^{N \times N}$, and assumed that the encoder network has $L$ layers, the node representation in the $i$-th layer of the encoder $\mathbf{H}^{(i)}\in \mathbb{R}^{d^{(i)} \times N}$ can be formulated as

\begin{equation}
\label{EQ1}
\mathbf{H}^{(i)}=g(\mathbf{H}^{(i-1)},\mathbf{A}^{(i-1)}|\mathbf{\Omega}^{(i-1)}),
\end{equation}
where $\mathbf{\Omega}^{(i)}$ is the trainable parameters in the encoder, and $g(\cdot)$ is an nonlinear mapping function which represents the encoding process with arbitrary GNN method. In this paper, we employ a two-layer GAT network as the nonlinear mapping encoder. Specifically, the node attribute is considered as the initial node representation, i.e., $\mathbf{H}^{(0)}=\mathbf{X}$, and the output of the encoder is treated as the node representation in the $L$-th layer of the encoder, i.e., $\mathbf{Z}=\mathbf{H}^{(L)}$.

The decoder is just reversing the encoding process, which means each decoder layer attempts to reverse the process of its corresponding encoder layer, and in this way, the node representation can be obtained without supervision. Given the node representation $\mathbf{Z}\in \mathbb{R}^{d^{(L)} \times N}$ and node relation matrix in the $L$-th layer $\mathbf{A}^{(L)} \in \mathbb{R}^{N \times N}$, and assumed that the decoder network has $L$ layers, the node representation in the $i$-th layer of the decoder $\hat{\mathbf{H}}^{(i)}\in \mathbb{R}^{d^{(i)} \times N}$ can be formulated as

\begin{equation}
\label{EQ2}
\hat{\mathbf{H}}^{(i)}=\hat{g}(\hat{\mathbf{H}}^{(i-1)},\hat{\mathbf{A}}^{(i-1)}|\hat{\mathbf{\Omega}}^{(i-1)}),
\end{equation}
where $\hat{\mathbf{\Omega}}^{(i)}$ is the trainable parameter in the decoder and $\hat{g}(\cdot)$ is an nonlinear mapping function in the decoder process. To be consistent with the encoder, the decoder adopts GAT as the nonlinear mapping function. Specifically, the input of the decoder is the node representation after self-expression layer transformation, i.e., $\hat{\mathbf{H}}^{(L)}=\mathbf{ZC}$, where $\mathbf{C}\in \mathbb{R}^{N \times N} $ is the self-expression coefficient matrix, and the output of the decoder is treated as the reconstructed node attribute, i.e., $\hat{\mathbf{X}}=\hat{\mathbf{H}}^{(0)}$.

To make sure the node representation can learn more favorable and satisfactory information from both node attributes and graph structures, the node representation reconstruction loss $\mathcal{L}_{rec}$ is defined as

\begin{equation}
\label{EQ3}
\mathcal{L}_{rec}=\underset{\mathbf{\Omega}}\min \frac{1}{2} \|\mathbf{X}-\hat{\mathbf{X}}\|_{F}^{2}.
\end{equation}
where $\mathbf{\Omega}$ is the trainable parameter in the Symmetric Feature Extraction Module.

\subsection{Neighborhood Contrast Module}
Now, we have the node representation from the Symmetric Feature Extraction Module, however, it is not just designed for the graph clustering task. To sufficiently improve the quality of the node representation obtained from the Symmetric Feature Extraction Module, we propose a Neighborhood Contrast Module.

The goal of the attributed graph clustering is to divide the nodes with similar attributes into the same cluster, and the nodes in different clusters have low similarity to each other. As aforementioned, we can leverage the contrastive learning method, which has been widely adopted in the representation learning field to 'push closer’ the given node representation and similar node representations, meanwhile, 'push away' it from others.

Then the problem lies in how to distinguish whether the node representation is similar to the given or not, in other words, how to define the positive/negative pairs in the contrastive learning process. As the previous works \cite{zhu2021graph, you2020graph, zhu2020deep, xia2021self, DBLP:conf/nips/YouCSCWS20} illustrated, graph contrastive learning is executed at instance-level and positive/negative pairs are constructed by graph data augmentation. However, it is still uncertain what kind of graph data augmentation is suitable for the corresponding clustering tasks since inappropriate augmentation may limit the performance. To get rid of graph data augmentation and consider the node similarity can be described by the distance of their features, in this paper, we obtain the top K nearest neighborhood of the given node by the K-Nearest Neighbor (KNN) method and treat them as positive pairs.

\begin{figure}[h]
\centering
\includegraphics[width=3.5in]{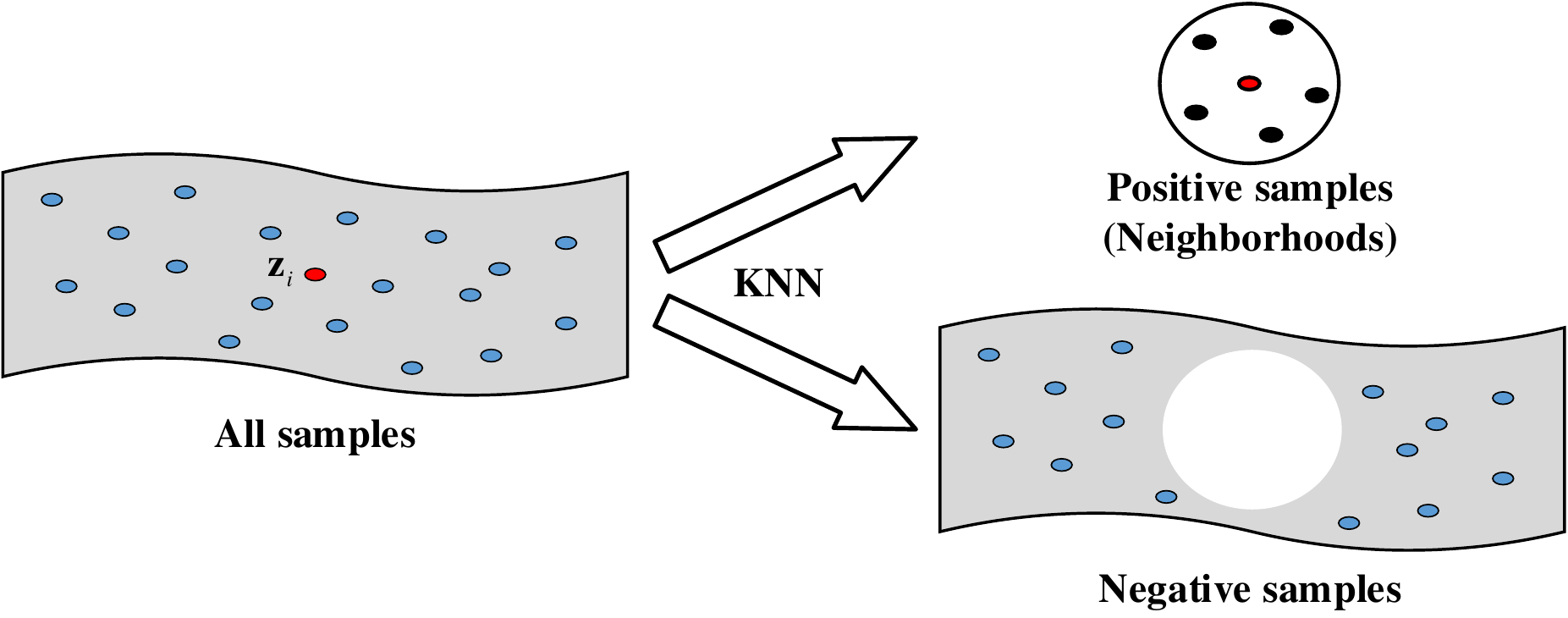}
\caption{The selection of positive and negative samples in Contrastive Neighborhood Module.}
\label{fig2}
\end{figure}

To be specific, for the given node representation $\mathbf{z}_{(i)}\in \mathbb{R}^{d^{(i)}}$, i.e., the red node in $N$ node samples $\{\mathbf{z}_1, \mathbf{z}_2, ..., \mathbf{z}_N\}$ and given neighborhood size $K$ in Fig. \ref{fig2}, we can calculate top $K$ similar nodes $\tilde{\mathbf{z}}_{j} \in \mathbb{R}^{d^{(L)}}$ of the given node $\mathbf{z}_{(i)}$ in terms of KNN method and regard them as positive samples, leaving other $N$-$K$-1 samples to be negative. After that, each positive sample can form a positive pair with the given sample $\mathbf{z}_{i}$, while each negative sample can form a negative pair with $\mathbf{z}_{i}$.

The neighborhood contrast loss $\ell_{i}$ for a single node $\mathbf{z}_{i}$ is in the form of

\begin{equation}
\label{EQ4}
\ell_{i}=\underset{\mathbf{\Omega}}\min-\log \frac{\sum\nolimits_{j=1}^{K}\exp(s(\mathbf{z}_i,\tilde{\mathbf{z}}_j))}{\sum\nolimits_{p\neq i}^{N}\exp(s(\mathbf{z}_i,\mathbf{z}_p))},
\end{equation}
where $\mathbf{\Omega}$ is the trainable parameter in the Symmetric Feature Extraction Module, $K$ is the pre-defined neighborhood size, $\tilde{\mathbf{z}}_j$ is the neighborhood of $\mathbf{z}_i$ calculated by KNN, $\sum\nolimits_{j=1}^{K}\exp(s(\mathbf{z}_i,\tilde{\mathbf{z}}_j))$ denotes the total similarity of positive pairs, and the pair-wise similarity $s(\mathbf{a},\mathbf{b})$ is measured by cosine distance, i.e.,

\begin{equation}
\label{EQ5}
s(\mathbf{a},\mathbf{b})=\frac{(\mathbf{a})(\mathbf{b})^{\top}}{\|\mathbf{a}\| \|\mathbf{b}\|},
\end{equation}
where $\mathbf{a}$ and $\mathbf{b}$ are two arbitrary vectors.

Taking all the nodes in graph $\mathcal{G}$ into account, we have the following neighborhood contrast loss $\mathcal{L}_{nbr}$, i.e,

\begin{equation}
\label{EQ6}
\mathcal{L}_{nbr}=\underset{\mathbf{\Omega}}\min\sum_{i=1}^{N}\ell_{i}.
\end{equation}

\subsection{Contrastive Self-Expression Module}
Despite the aforementioned modules could output a high-quality node representation $\mathbf{Z}$, there is no guarantee that it can be well used to construct the self-expression coefficient matrix $\mathbf{C}$ in the deep subspace clustering process. Therefore, we propose the Contrastive Self-Expression Module for learning a more discriminative self-expression coefficient matrix.

The deep subspace clustering method learn the self-expressiveness by introducing a self-expression layer, i.e., a fully connected layer without a bias between encoder and decoder. Specifically, for a given node representation $\mathbf{z}_i$, the self-expression layer can calculate the linear combination of other nodes $\mathbf{z}_{j,j\neq i}$ to express the reconstructed node representation $\hat{\mathbf{z}}_{i}$ as

\begin{equation}
\label{EQ7}
\hat{\mathbf{z}}_{i}=\sum_{i\neq j}c_{ij}\mathbf{z}_j.
\end{equation}

Existing deep subspace clustering based methods simply optimize $\mathbf{C}$ by minimize the self-expression loss $\mathcal{L}_{se}$, i.e,

\begin{equation}
\label{EQ8}
\mathcal{L}_{se}=\underset{\mathbf{\Omega}}\min \|\mathbf{Z}-\mathbf{ZC}\|_{F}^{2},
\end{equation}
where $\mathbf{C}\in \mathbb{R}^{N \times N}$ is the self-expression coefficient matrix which describes the self-expressiveness and consists of each linear combination coefficient $c_{ij}$.

However, the conventional self-expression loss $\mathcal{L}_{se}$ only consider the corresponding node representation before and after the reconstruction of the self-expression layer, i.e., $\mathbf{z}_i$ and $\hat{\mathbf{z}}_i$, neglecting the information of other nodes, which degrades the learning quality of $\mathbf{C}$ and limits the clustering performance. To this end, we propose the Contrastive Self-Expression Module which adopts the contrastive learning method by 'push closer' the given node representation $\mathbf{z}_i$ and the corresponding node representation $\hat{\mathbf{z}}_i$, and 'push away' it from other nodes. To our knowledge, this could be the first attempt which improves the performance of subspace clustering with contrastive learning strategy.

\begin{figure}[h]
\centering
\includegraphics[width=3.5in]{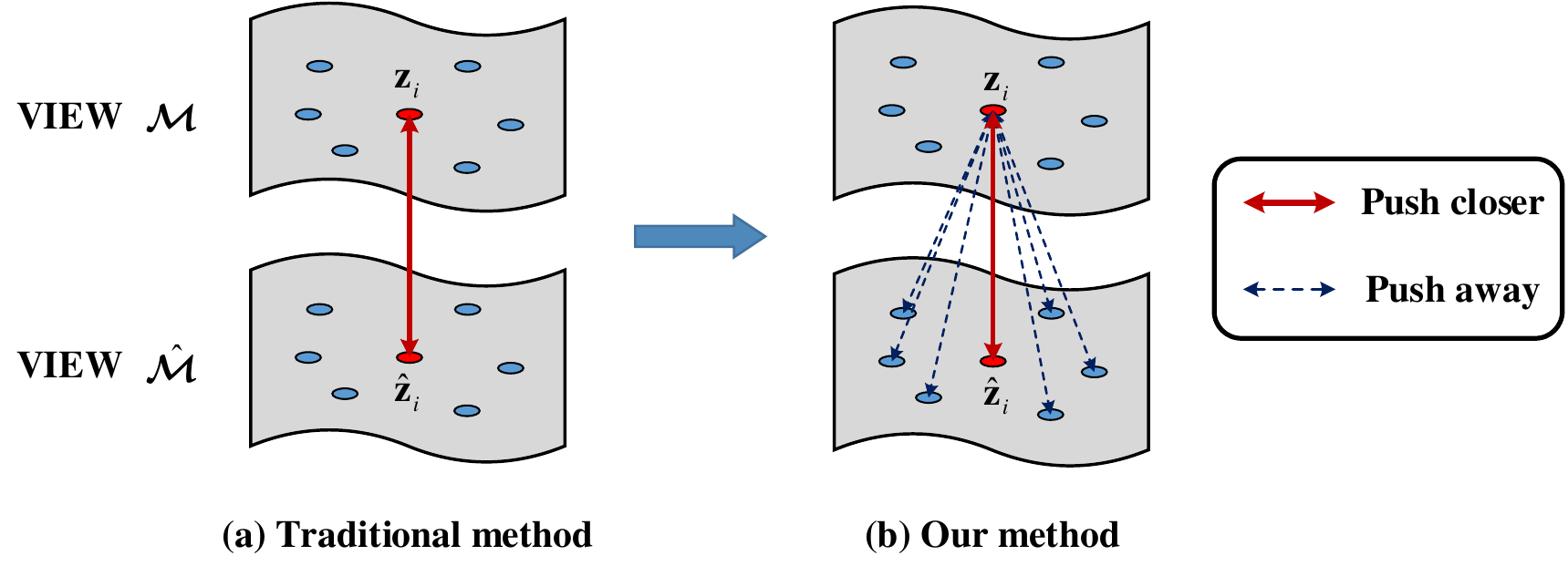}
\caption{The contrastive method in self-expression layer.}
\label{fig3}
\end{figure}

Specifically, as shown in Fig. \ref{fig3}, we have the attributed graph views before and after the self-expression layer as $\mathcal{M}$ and $\hat{\mathcal{M}}$, respectively. For a given node $\mathbf{z}_i$ in $\mathcal{M}$, it can form a positive pair with the corresponding node $\hat{\mathbf{z}}_i$ in $\hat{\mathcal{M}}$, leaving other $N$-1 nodes in $\hat{\mathcal{M}}$ to be negative pairs. The contrastive self-expression loss $\ell_{i}$ for a single node $\mathbf{z}_i$ is in the form of

\begin{equation}
\label{EQ9}
\ell_{i}=\underset{\mathbf{\Omega},\mathbf{C}}\min-\log \frac{\exp(s(\mathbf{z}_i,\hat{\mathbf{z}}_i))}{\sum\nolimits_{j=1}^{N}\exp(s(\mathbf{z}_i,\hat{\mathbf{z}}_j))},
\end{equation}
where $\hat{\mathbf{z}}_{i}=\sum_{i\neq j}c_{ij}\mathbf{z}_j$ denotes the reconstructed node representation in $\hat{\mathcal{M}}$, $s(\mathbf{a},\mathbf{b})$ is the pair-wise similarity.

Taking all the nodes in graph $\mathcal{G}$ into account, we have the proposed contrastive self-expression loss $\mathcal{L}_{cse}$, i.e,

\begin{equation}
\label{EQ10}
\mathcal{L}_{cse}=\underset{\mathbf{\Omega},\mathbf{C}}\min\sum_{i=1}^{N}\ell_{i}.
\end{equation}

As deep subspace clustering method\cite{ji2017deep}, the self-expression coefficient matrix $\mathbf{C}$ is leveraged to construct the affinity matrix $\mathbf{\Lambda}$ for spectral clustering. Mathematically, we can have the following regularization term $\mathcal{L}_{coef}$, i.e,

\begin{equation}
\label{EQ11}
\mathcal{L}_{coef}=\underset{\mathbf{C}}\min\|\mathbf{C}\|_p,
\end{equation}
where $\|\cdot\|_p$ denotes an arbitrary matrix norm, in this paper, we adopt $\ell_2$ norm.

\subsection{Optimize}
Combining Eq. (\ref{EQ3}), Eq. (\ref{EQ6}), Eq. (\ref{EQ10}) and Eq. (\ref{EQ11}), NCAGC optimizes the following loss function
\begin{equation}
\label{EQ12}
\mathcal{L}_{total}=\underset{\mathbf{\Omega},\mathbf{C}}\min\mathcal{L}_{rec}+\lambda_1\cdot\mathcal{L}_{nbr}+\lambda_2\cdot\mathcal{L}_{cse}+\lambda_3\cdot\mathcal{L}_{coef},
\end{equation}
where $\lambda_1$, $\lambda_2$, $\lambda_3$ are three trade-off parameters which keep the balance of different loss function. By optimizing Eq.(\ref{EQ12}), the node representation will be extracted with neighborhood information, meanwhile, the attributed graph clustering and node representation learning process are seamlessly connected. Once the network is trained, we can use the optimized $\mathbf{C}$ to construct $\mathbf{\Lambda}$ for spectral clustering. In this paper, we adopt the affinity matrix constructing method in EDSC\cite{ji2014efficient}. Algorithm \ref{alg:algorithm} shows the optimizing process of the proposed NCAGC.

\begin{algorithm}[h]
\caption{NCAGC}
\label{alg:algorithm}
\textbf{Input}: Graph dataset $\mathcal{G}$ with feature matrix $\mathbf{X}$ and adjacency matrix $\mathbf{A}$.\\
\textbf{Parameter}: Trade-off parameters $\lambda_1$, $\lambda_2$, $\lambda_3$, neighborhood size $K$, cluster numbers $k$, training epoches $T$, iterator $t$, learning rate $lr$.
\begin{algorithmic}[1] 
\STATE Initialize the self-expression matrix $\mathbf{C}$ with $1\times10^{-4}$;
\WHILE{$t < T$}
\STATE Calculate the neighborhood nodes by KNN;
\STATE Select positive pairs in the overall framework;
\STATE Joint train the overall framework by Eq. (\ref{EQ12});
\STATE Update $\mathbf{C}$ from the optimized network;
\STATE Obtain affinity matrix $\mathbf{\Lambda}$ using $\mathbf{C}$;
\STATE Spectral clustering on $\mathbf{\Lambda}$;
\ENDWHILE
\end{algorithmic}
\textbf{Output}: Clustering label $\mathbf{Y}$.
\end{algorithm}

\section{Experiment}
In this section, we conduct extensive experiments to fully analyze the effectiveness of our proposed NCAGC and the proposed Neighborhood Contrast Module and Contrastive Self-Expression Module.

\subsection{Benchmark Datasets}
To evaluate the performance of our proposed NCAGC, we conduct the following experiment on four attributed graph datasets (Cora, Citeseer, Wiki and ACM).

To be specific, Cora\cite{mccallum2000automating} and Citeseer\cite{giles1998citeseer} are citation networks where nodes denote publications and are connected if cited by another. The node attribute of Cora and Citeseer datasets is described by a spare bag of word feature vectors. Wiki\cite{yang2015network} is a webpage network where nodes denote webpages and are connected if linked by another. The node attribute of Wiki dataset is described by TF-IDF weighted vectors. ACM\cite{fan2020one2multi} is a paper network where nodes denote papers and they are connected if they are written by the same author. The node attribute of ACM dataset is bag of words of keywords. The details and statistics of these datasets are shown in Table \ref{tab:table1}.

\begin{table*}[t]
\caption{The statistics of the datasets\label{tab:table1}}
\centering
\begin{tabular}{l|c|c|c|c|c|c}
\hline
\bf{Datasets} & \bf{Nodes} & \bf{Relation Types} & \bf{Edges} & \bf{Feature Types} & \bf{Attributes} & \bf{Classes}\\
\hline
Cora & 2,708 & Citation Network & 5,429 & Bag of words of keywords & 1,433 & 7\\
Citeseer & 3,327 & Citation Network & 4,732 & Bag of words of keywords & 3,703 & 6\\
Wiki & 2,405 & Webpage Network & 17,981 & TF-IDF & 4,973 & 17\\
ACM & 3,025 & Paper Network & 29,281 & Bag of words of keywords & 1,870 & 3\\
\hline
\end{tabular}
\end{table*}

\subsection{Clustering Metrics}
Similar to the previous work\cite{wang2021multi}, we leverage three widely-used clustering metrics to evaluate the generic and effectiveness of our proposed NCAGC, namely Accuracy (ACC), Normalized Mutual Information (NMI), and Adjusted Rand Index (ARI). For all these metrics, a better score denotes a better clustering performance.

\subsection{Baseline Methods}
To better assess the clustering performance of our proposed NCAGC over the aforementioned datasets, we choose 16 state-of-the-art basis methods as competitors. They can be generally divided into the following three categories:

{\bf 1) Method only using node attributes}, which adopts the raw node attribute as input.
\begin{itemize}{}{}
\item {K-means\cite{hartigan1979algorithm} is a classical shallow clustering method, which alternately updates the location of the cluster center and the distance of the sample to the cluster center.}
\end{itemize}

{\bf 2) Method only using graph structures}, which adopts the raw topological graph structure as input.
\begin{itemize}{}{}
\item{Spectral\cite{ng2001spectral} is a classical shallow clustering method based on graph theory, which takes the node adjacency matrix as the similarity matrix.}
\end{itemize}

{\bf 3) Method using both node attributes and graph structures}, which takes node attributes and graph structures as input.
\begin{itemize}{}{}
\item{GAE\cite{kipf2016variational} is an unsupervised graph embedding method with a GCN-based encoder and an inner product decoder.}
\item{VGAE\cite{kipf2016variational} is the variational version of GAE.}
\item{ARGAE\cite{pan2018adversarially} is an adversarial regularized unsupervised graph embedding method with generative adversarial methods.}
\item{ARVGAE\cite{pan2018adversarially} is the variational version of ARGAE.}
\item{MGAE\cite{wang2017mgae} is a marginalized unsupervised graph clustering method, which employs a mapping loss to optimize the unsupervised graph feature extraction process and adopts spectral clustering to obtain clustering labels.}
\item{GATE\cite{salehi2019graph} employs a node attribute reconstruct loss and graph structure reconstruct loss to optimize node embedding.}
\item{DAEGC\cite{wang2019attributed} is an end-to-end graph clustering method employing attention mechanism in node representation learning and adopts a clustering loss for node clustering.}
\item{SDCN\cite{bo2020structural} is an end-to-end graph clustering method employing both auto-encoder and graph convolutional network during the node representation process.}
\item{DFCN\cite{tu2020deep} introduces an information fusion module to enhance the performance of SDCN.}
\item{AGC\cite{zhang2019attributed} utilizes adaptive graph convolution layers for different datasets.}
\item{GALA\cite{park2019symmetric} proposes a symmetric graph convolutional autoencoder for graph clustering, which produces a low-dimensional graph representation for clustering.}
\item{MSGA\cite{wang2021multi} introduces multi-scale self-expression layers and self-supervised method improving the clustering performance.}
\item{GRACE\cite{zhu2020deep} is an unsupervised graph embedding method with graph contrastive learning, which generates two views for comparison using graph augmentation.}
\item{GCA\cite{zhu2021graph} introduces an adaptive graph data augmentation based on GRACE.}
\end{itemize}

\subsection{Parameter Setting}

\begin{table*}[t]
\caption{Parameter Settings on NCAGC\label{tab:table2}}
\centering
\begin{tabular}{l|c|c|c|c|c|c|c|c}
\hline
\bf{Datasets} & \bf{Learning rate} & \bf{Encoder dimension} & \bf{GNN} & \bf{Activation function} & \bf{Neighborhood size} & \bf{$\lambda_1$}& \bf{$\lambda_2$}& \bf{$\lambda_3$}\\
\hline
Cora & 0.0001 & 1433 - 1024 - 512	& GAT & prelu & 10 & 10	& 10 & 10\\
Citeseer & 0.0001 & 3327 - 1024 - 1024	& GAT & prelu & 10 & 100 & 1 & 1\\
Wiki & 0.0001 & 4973 - 1024 - 512 & GAT & prelu & 10 & 10 & 1 & 10\\
ACM & 0.0005 & 1870 - 1024 - 512 & GAT	& prelu	& 10 & 100 & 200 & 3,500\\
\hline
\end{tabular}
\end{table*}

\begin{table*}[t]
\caption{Clustering Performance of NCAGC on Graph Datasets\label{tab:table3}}
\centering
\begin{tabular}{l|c|c c c| c c c | c c c | c c c}
\hline
\multirow{2}{*}{\bf{Methods}} & \multirow{2}{*}{\bf{Input}}  & \multicolumn{3}{c|}{Cora} & \multicolumn{3}{c|}{Citeseer} & \multicolumn{3}{c|}{Wiki} & \multicolumn{3}{c}{ACM}\\
 & & ACC & NMI	& ARI &	ACC &	NMI &	ARI &	ACC &	NMI &	ARI &	ACC &	NMI &	ARI\\
\hline
K-Means & Feature & 0.492 &	0.321 &	0.230 & 0.540 &	0.305 &	0.278 &	0.417 &	0.440 &	0.150 &	0.673 &	0.324 &	0.306\\
\hline
Spectral &	Graph &	0.367 &	0.127 &	0.031 &	0.239 &	0.056 &	0.010 &	0.220 &	0.182 &	0.015 &	0.368 &	0.007 &	0.006\\
\hline
GAE &	Both &	0.502 &	0.329 &	0.217 &	0.412 &	0.183 &	0.189 &	0.173 &	0.119 &	0.160 &	0.845 &	0.553 &	0.594\\
VGAE &	Both &	0.559 &	0.384 &	0.254 &	0.447 &	0.260 &	0.205 &	0.286 &	0.302 &	0.263 &	0.847 &	0.556 &	0.601\\
MGAE &	Both &	0.684 &	0.511 &	0.444 &	0.660 &	0.412 &	0.413 &	0.514 &	0.485 &	0.350 &	0.881 &	0.621 &	0.687\\
ARGAE &	Both & 0.640 &	0.449 &	0.352 &	0.573 &	0.350 &	0.341 &	0.380 &	0.344 &	0.112 &	0.843 &	0.545 &	0.606\\
ARVGAE &Both &	0.638 &	0.450 &	0.374 &	0.544 &	0.261 &	0.245 &	0.386 &	0.338 &	0.107 &	0.845 &	0.547 &	0.603\\
GATE  & Both &  0.658 &	0.527 &	0.451 &	0.616 &	0.401 &	0.381 &	0.465 & 0.428 &	0.316 & 0.863 &	0.574 &	0.624\\
DAEGC &	Both &	0.704 & 0.528 &	0.496 & 0.672 &	0.397 &	0.410 &	0.482 &	0.448 &	0.330 &	0.874 &	0.591 &	0.633\\
SDCN &	Both &  0.712 & 0.535 & 0.506 &	0.659 &	0.387 &	0.401 &	0.385 &	0.375 &	0.285 &	0.904 &	0.683 &	0.731\\
DFCN &	Both &	0.740 &	0.561 &	0.527 &	0.695 &	0.439 &	\bf{0.455} &	0.416 &	0.400 &	0.332 &	0.909 &	\bf{0.694} & 0.749\\
AGC &	Both & 0.689 & 0.537 & 0.493 & 0.670 & 0.411 & 0.415 & 0.477 & 0.453 & 0.325 & 0.893 & 0.653 & 0.710\\
GALA &	Both &	0.746 &	0.577 &	\bf{0.532} &	0.693 &	\bf{0.441} &	0.446 &	\bf{0.525} &	0.480 &	\bf{0.366} &	0.908 &	0.688 &	0.747\\
MSGA & Both & \bf{0.747} & \bf{0.578}  & 0.519 &	\bf{0.698} &	0.415 & 0.433 &	0.522 &	\bf{0.481} &	0.323 &	\bf{0.911} &	0.692 &	\bf{0.752}\\
GRACE & Both &	0.671 & 0.530 & 0.485 &	0.651 &	0.395 & 0.402 &	0.475 &	0.430 &	0.308 &	0.881 &	0.651 &	0.695\\
GCA	& Both & 0.696 & 0.538 &	0.491 &	0.643 &	0.393 &	0.405 &	0.477 &	0.435 &	0.315 &	0.889 &	0.653 &	0.698\\
\hline
NCAGC  & Both &	\bf{0.764} & \bf{0.603} &	\bf{0.552} &	\bf{0.707} &	\bf{0.438} &	\bf{0.463} &	\bf{0.530} &	\bf{0.485} &	\bf{0.336} &	\bf{0.917} &	\bf{0.711} &	\bf{0.769}\\
\hline
\end{tabular}
\end{table*}

The proposed NCAGC and baseline methods are implemented on a machine equipped with an NVIDIA RTX 3080 GPU. The deep learning environment for our proposed NCAGC is Pytorch and PyTorch Geometric.

For all baseline methods, the parameters are set according to the corresponding papers for the best performance. For NCAGC, we employ a two-layer GAT network\cite{velivckovic2017graph} as encoder and a symmetrical decoder with Adam optimizer. The learning rate $lr$ is set as $0.0001$ on Cora, Citeseer and Wiki and $0.0005$ on ACM. The dimension of the latent layer for the encoder is searched in the range of $\{256, 512, 1024, 2048\}$. Specifically, the dimension is set as $[1024, 512]$ on Cora, $[1024, 1024]$ on Citeseer, $[1024, 512]$ on Wiki and $[1024, 512]$ on ACM, respectively. We use prelu as activation function and the number of neighbors is set as $10$. To obtain a balance of different loss functions and get optimal results, we turn the trade-off parameters $\lambda_1$, $\lambda_2$, $\lambda_3$ in the range of $\{0.001, 0.01, 0.1, 1, 10, 100, 1000\}$. Specially, the trade-off parameters $\lambda_1$, $\lambda_2$, $\lambda_3$ are set as $[10, 10, 10]$ on Cora, $[10, 1, 1]$ on Citeseer,  $[10, 1, 10]$ on Wiki and $[100, 200, 3500]$ on ACM, respectively. We train NCAGC for $400$ iterations on Cora, $200$ iterations on Citeseer and ACM and $300$ iterations on Wiki, respectively. The detailed parameter settings on NCAGC is shown in Table \ref{tab:table2}.

\subsection{Experiment Results}
To evaluate the performance of NCAGC, we run the aforementioned 16 methods 10 times and report the average score to avoid randomness. The clustering performance is summarized in Table \ref{tab:table3}, where the top 2 scores are represented by {\bf bold} value.

As shown in Table \ref{tab:table3}, it can be observed that our proposed NCAGC achieves competitive performance compared with all the baseline methods according to three clustering metrics, which demonstrates the effectiveness of our proposed method. To be specific, we can have the following interesting observations:

\begin{itemize}
\item{The proposed NCAGC and other GCN based methods (GAE, ARGAE, GATE, DAEGC, SDCN, DFCN, AGC, GALA, MSGA, GRACE, and GCA) show superiority over K-Means and Spectral method, which demonstrates that the method with both node attributes and graph structures performs better than only using one of them. This is because GCN based method can exploit node attributes and graph structures simultaneously with a multi-layer nonlinear network.}
\item{The proposed NCAGC achieves better clustering performance than some representative graph auto-encoder based methods (GAE, ARGAE, and GATE). Especially, for Cora dataset, NCAGC surpasses GAE by about 51.9\%, 83.3\% and 154.3\% and GATE by about 15.9\%, 14.4\% and 22.3\% on ACC, NMI and ARI, respectively. This result indicates the contrastive strategy of the Neighborhood Contrast Module employed by considering 'push closer' neighbor nodes, i.e., positive pairs, can help to improve the learned node representation quality.}
\item{In some cases, the clustering performance of the method using graph contrastive learning (GRACE and GCA) is inferior to the clustering-oriented method (DAEGC, SDCN, DFCN, AGC, GALA, MSGA, and NCAGC). This can be concluded that NCAGC integrates node representation learning and clustering into a unified framework, which greatly helps the network learn a clustering-oriented representation. In contrast, GRACE and GCA perform node representation learning and clustering in two separate steps, which limited the performance.}
\item{The proposed NCAGC significantly outperforms all the methods in terms of ACC metric and achieves the top 2 score in terms of NMI and ARI metrics. Taking the clustering performance on Cora dataset as an example, NCAGC surpasses the sub-optimal method by 2.1\%, 4.3\% and 3.7\% in terms of ACC, NMI and ARI, respectively. The remarkable result verifies the powerful clustering ability of NCAGC, which benefits from the incorporation of the Neighborhood Contrast Module and the Contrastive Self-Expression Module.}
\end{itemize}

\begin{figure*}[t]
\centering
\subfloat[]{\includegraphics[width=2.4in]{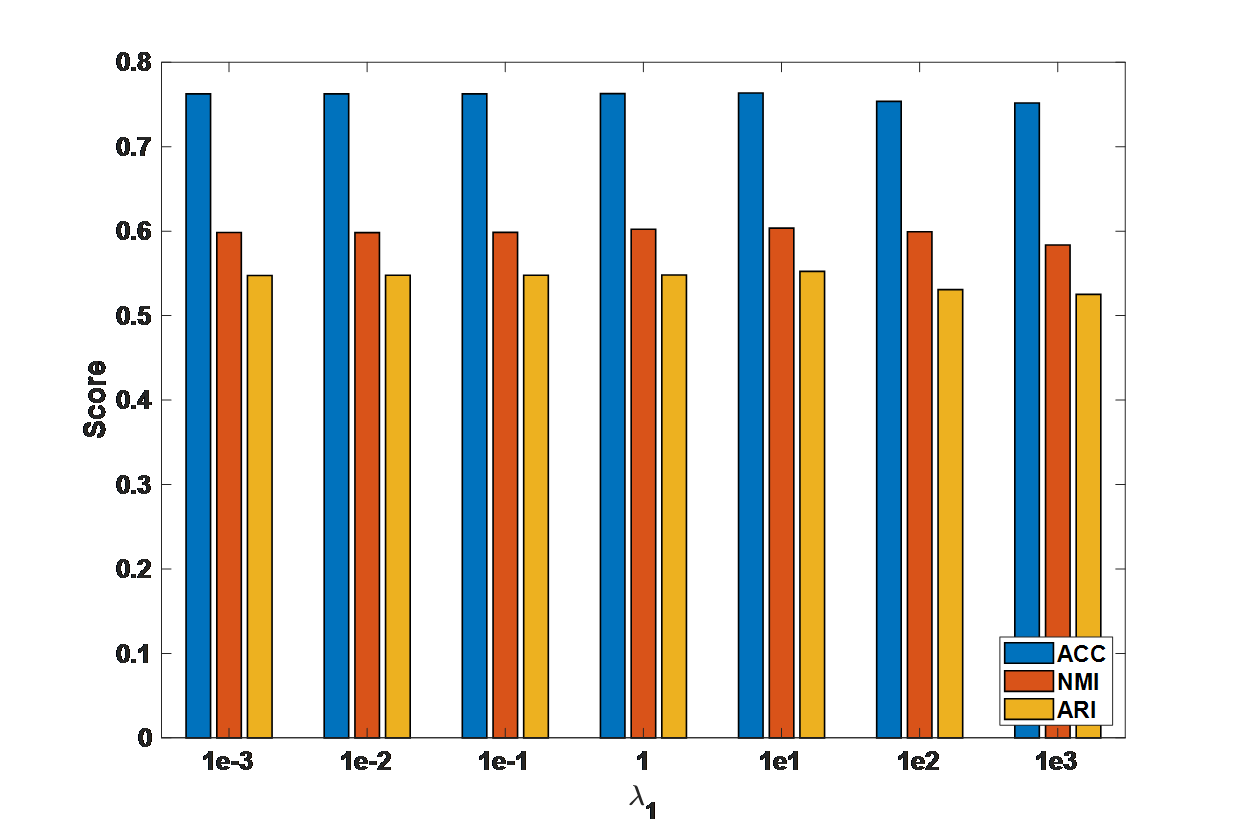}%
\label{fig4_first_case}}
\subfloat[]{\includegraphics[width=2.4in]{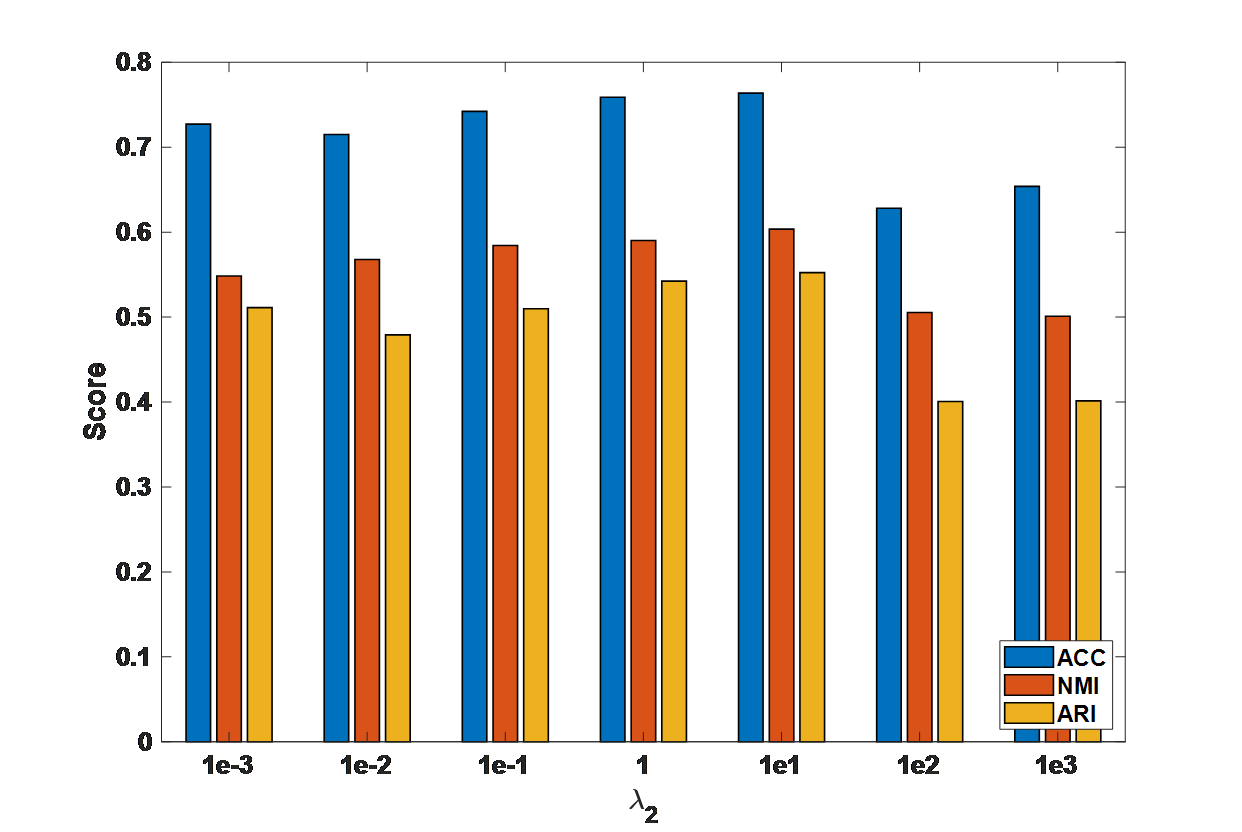}%
\label{fig4_second_case}}
\subfloat[]{\includegraphics[width=2.4in]{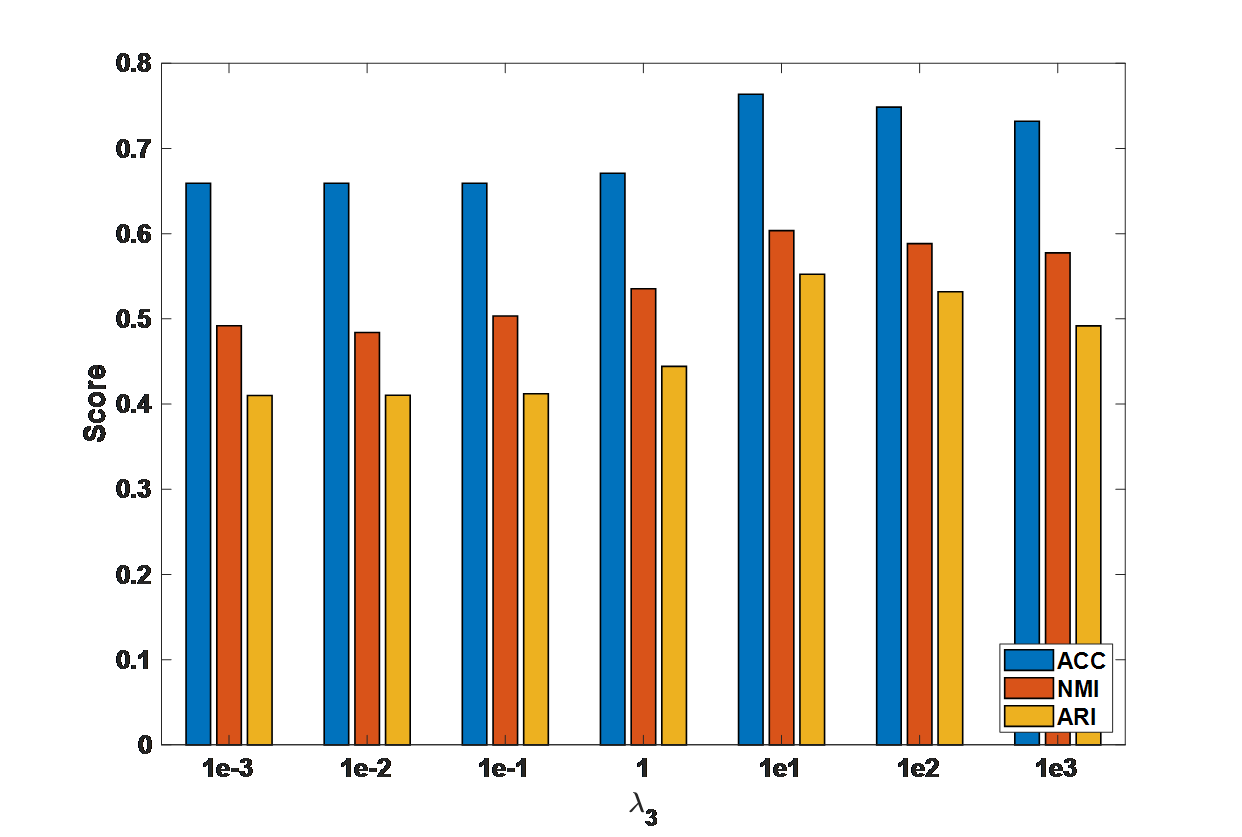}%
\label{fig4_third_case}}
\caption{Influence of individual values on ACC, NMI and ARI of Cora dataset.}
\label{fig4}
\end{figure*}

\begin{figure*}[t]
\centering
\subfloat[]{\includegraphics[width=2.4in]{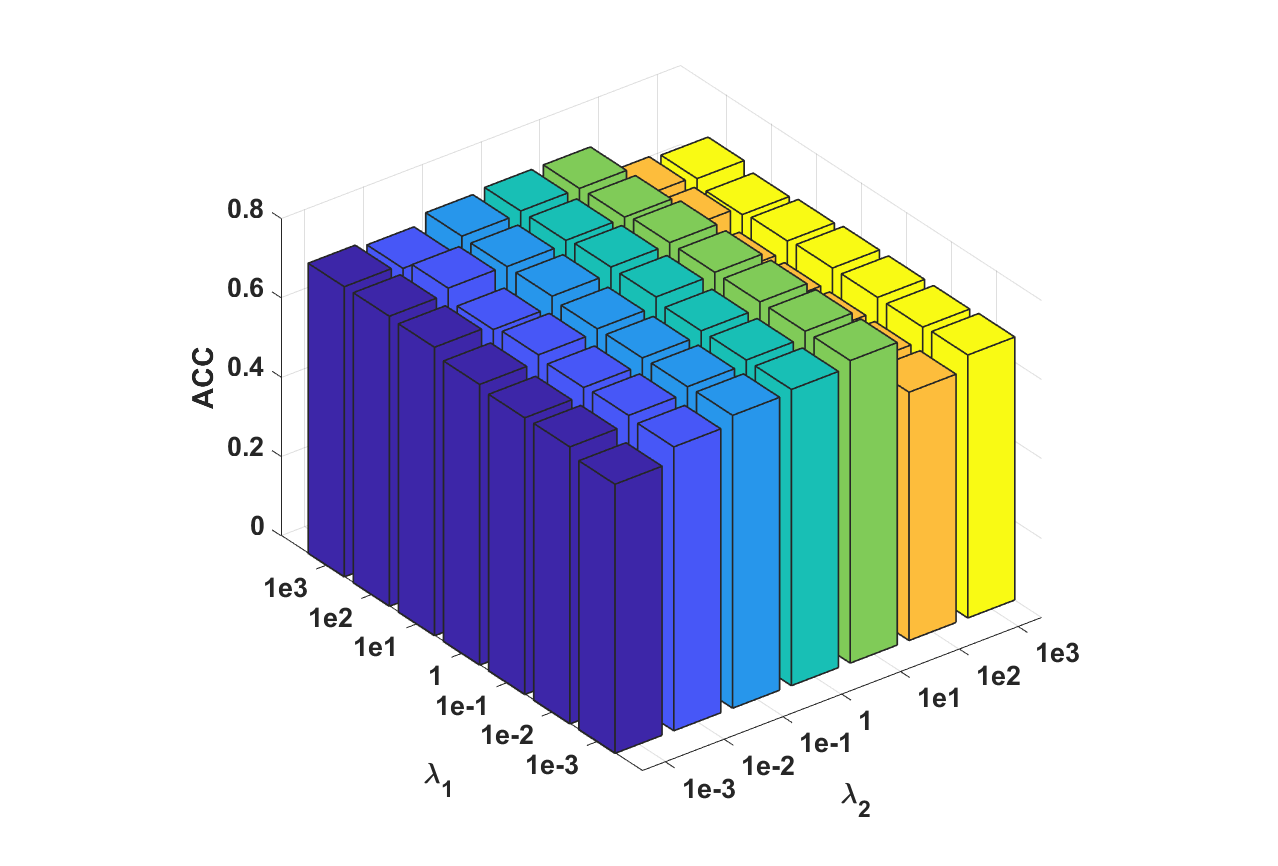}%
\label{fig5_first_case}}
\subfloat[]{\includegraphics[width=2.4in]{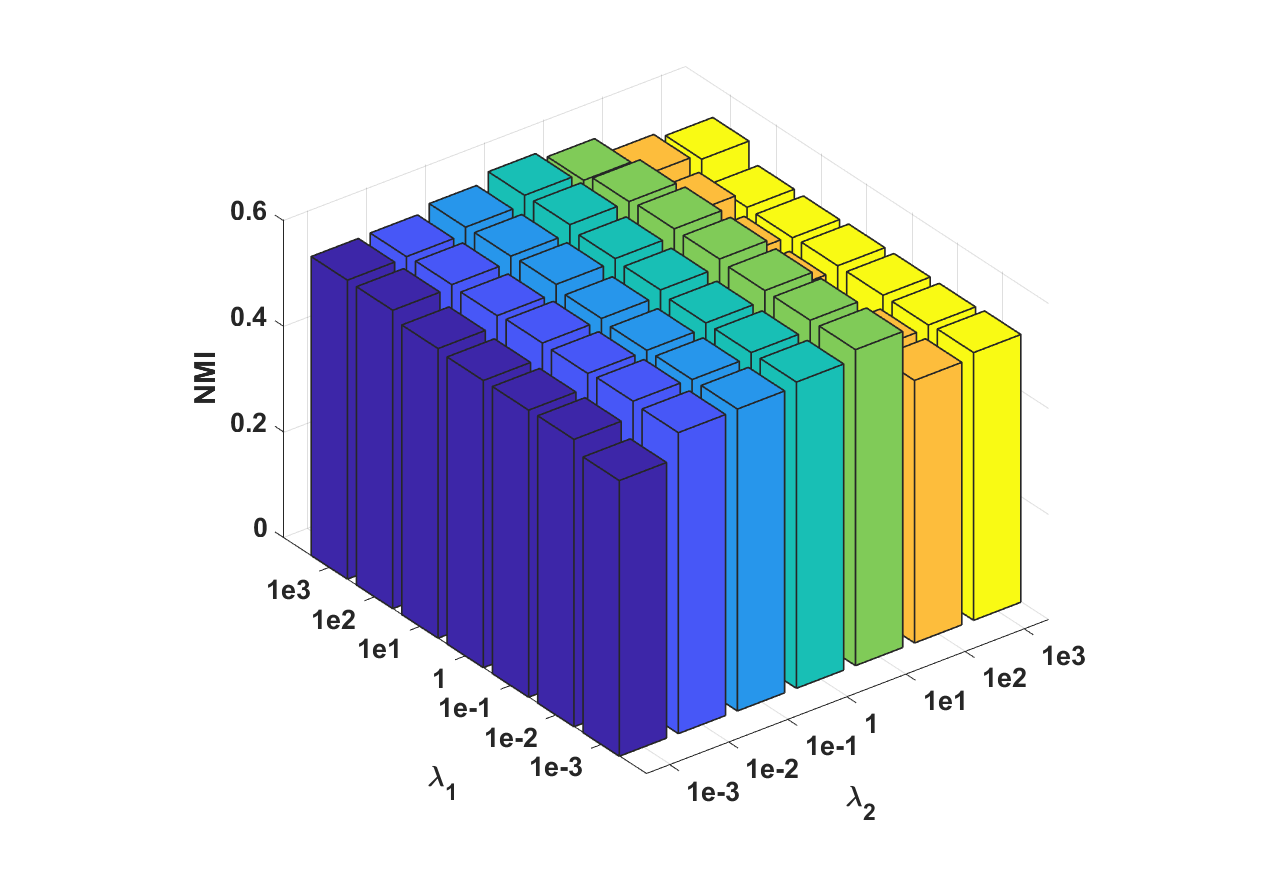}%
\label{fig5_second_case}}
\subfloat[]{\includegraphics[width=2.4in]{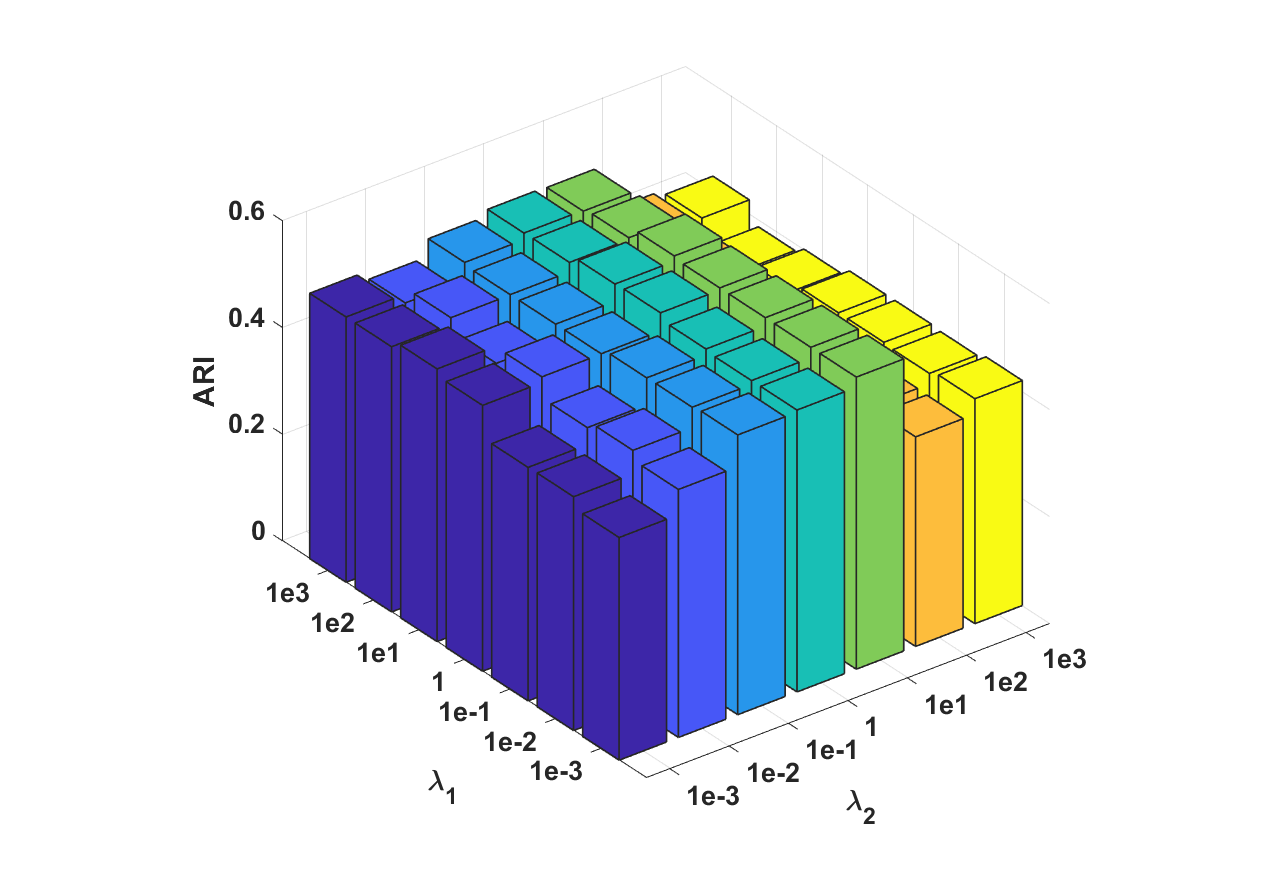}%
\label{fig5_third_case}}
\caption{Influence of $\lambda_1$ and $\lambda_2$ values on ACC, NMI and ARI of Cora dataset.}
\label{fig5}
\end{figure*}

\subsection{Parameter Analysis}

There are three trade-off parameters in our proposed NCAGC, namely $\lambda_1$, $\lambda_2$ and $\lambda_3$, which denote the weight of neighborhood contrast loss $\mathcal{L}_{nbr}$, contrastive self-expression loss $\mathcal{L}_{cse}$ and regularization of self-expression coefficient $\mathcal{L}_{coef}$, respectively. To this end, we take Cora dataset as an example and conduct the experiment to verify the effectiveness of different loss functions of NCAGC, besides, we also investigate the sensitivity of neighborhood size $K$ in the Neighborhood Contrast Module.

{\it 1) The effect of trade-off parameters:} We first analyze the influence of individual trade-off parameters via tuning the under-tested parameter from $10^{-3}$ to $10^{3}$ and fixing other parameters as in Table \ref{tab:table2}. Fig. \ref{fig4} shows the changing of the clustering performance by individual parameters. We notice that the score of clustering metrics keeps stable as $\lambda_1$ varied. It could be inferred that the contrastive method of the Neighborhood Contrast Module which 'push closer' the node representation of neighbors has strong robustness. Furthermore, we can observe that big $\lambda_2$ and small $\lambda_3$ will degrade the clustering performance, since $\lambda_2$ and $\lambda_3$ are related to the learning of the self-expression coefficient which is crucial to clustering. The result shows the self-expression layer of NCAGC could learn a more discriminative self-expression coefficient when $\lambda_2$ is under $10$ or over $10$.

Besides, to further analyze the effect of a comprehensive combination of these trade-off parameters, we vary the under-tested parameter from $10^{-3}$ to $10^{3}$ at the same time. From Fig. \ref{fig5}, we can observe that the clustering performance is slightly improved as $\lambda_2$ varies from $10^{-3}$ to $10^{3}$ and dramatically declined since $\lambda_2$ increases to $10^{3}$. The result can be inferred that the contrastive self-expression loss $\mathcal{L}_{cse}$ becomes very large as $\lambda_2$ varies which makes the overall loss function unbalanced and then degrades the performance.

{\it 2) The effect of neighborhood size:} To further explore the effectiveness and robustness of our proposed Neighborhood Contrast Module, we vary the size of node neighborhood $K$ in the range of $\{3, 5, 7, 10, 15, 20, 30\}$ while other parameters unchanged and report the clustering result of three datasets in Table \ref{tab:table4}. We can observe that the clustering results in terms of ACC, NMI and ARI fluctuated by 0.4\%, 0.8\% and 1.2\% on Cora dataset, 0.4\%, 0.7\% and 1.1\% on Citeseer dataset and 0.8\%, 3.1\% and 2.3\% on ACM dataset. As reported in Table \ref{tab:table4}, we find NCAGC achieves the best clustering performance when $K$=10 over several datasets. Besides, this can prove that our proposed Neighborhood Contrast Module can steadily improve performance.

\begin{table*}[t]
\caption{Parameter analysis on the size of neighborhood\label{tab:table4}}
\centering
\begin{tabular}{l|c c c|c c c|c c c}
\hline
\multirow{2}{*}{\bf{Size of $K$}} & \multicolumn{3}{c|}{Cora} & \multicolumn{3}{c|}{Citeseer} & \multicolumn{3}{c}{ACM}\\
&	ACC&	NMI&	ARI&	ACC&	NMI&	ARI&	ACC&	NMI&	ARI\\
\hline
$K$ = 3 &	0.762 & 0.601 & 0.550 &	0.704 & 0.433 & 0.457 & 0.912 & 0.697 & 0.756\\
$K$ = 5 & 0.761&	0.598 &	0.547 &	0.705 &	0.434 &	0.458 &	0.911 &	0.692 &	0.753\\
$K$ = 7&	0.762 &	0.597 &	0.550 &	0.706 &	0.437 &	0.460 &	0.911 &	0.694 &	0.755\\
$K$ = 10& 0.764 & 0.603 & 0.552 & 0.707 & 0.437 & 0.462 & 0.917 & 0.711 &	0.769\\
$K$ = 15& 0.763 & 0.601 & 0.552 & 0.706 & 0.436 & 0.461 & 0.909 & 0.690 &	0.750\\
$K$ = 20& 0.761 & 0.602 & 0.546 & 0.704 & 0.437 & 0.460 & 0.916 & 0.702 &	0.766\\
$K$ = 30& 0.762 & 0.602 & 0.550 & 0.704 & 0.434 & 0.458 & 0.909 & 0.690 &	0.749\\
\hline
\end{tabular}
\end{table*}

\begin{table*}[t]
\caption{Ablation Study of the model\label{tab:table5}}
\centering
\begin{tabular}{l|c c c|c c c|c c c}
\hline
\multirow{2}{*}{\bf{Strategies}} & \multicolumn{3}{c|}{Cora} & \multicolumn{3}{c|}{Citeseer} & \multicolumn{3}{c}{ACM}\\
&	ACC&	NMI&	ARI&	ACC&	NMI&	ARI&	ACC&	NMI&	ARI\\
\hline
NCAGC w/o NBR & 0.742 &	0.578 &	0.517 &	0.683 &	0.417 &	0.432 &	0.902 &	0.682 &	0.747\\
NCAGC w/o CSE &	0.685 & 0.546 & 0.469 & 0.643 & 0.395 & 0.380 & 0.730 &	0.400 &	0.353\\
NCAGC w/o ATT & 0.736 & 0.560 & 0.501 &0.703 & 0.426 & 0.455 & 0.907	& 0.681	& 0.745\\
\hline
NCAGC & 0.764 &	0.603 &	0.552 &	0.707 &	0.438 &	0.463 & 0.917 & 0.711 & 0.769\\
\hline
\end{tabular}
\end{table*}

\subsection{Ablation Study}
To validate the effectiveness of different strategies in NCAGC, we carry out three ablation studies on Cora, Citeseer and Wiki datasets.

{\it 1) Ablation study on Neighborhood Contrast Module:} Particularly, we compare NCAGC and NCAGC without using Neighborhood Contrast Module ({\bf termed NCAGC w/o NBR}) by removing neighborhood contrast loss $\mathcal{L}_{nbr}$. In this scenario, the total loss function of NCAGC w/o NBR can be described by $\mathcal{L}_{rec}+\mathcal{L}_{cse}+\mathcal{L}_{coef}$, where $\mathcal{L}_{rec}$ is the loss of node representation reconstruction, $\mathcal{L}_{cse}$ is the loss of self-expression employed instance contrastive learning method and $\mathcal{L}_{coef}$ is the regularization term. The result of the ablation study on the Neighborhood Contrast Module is shown in Table \ref{tab:table5}, comparing with NCAGC w/o NBR, we can observe that NCAGC outperforms Cora, Citeseer and ACM datasets by 2.8\%, 3.5\% and 1.6\% according to ACC, by 4.3\%, 5.0\% and 4.3\% according to NMI and by 6.7\%, 7.2\% and 2.9\% according to ARI. This is because using the Neighborhood Contrast Module can better extract node representation and help to improve the clustering performance.

{\it 2) Ablation study on Contrastive Self-Expression Module:} Furthermore, we evaluate the effectiveness of Contrastive Self-Expression Module by using self-expression loss $\mathcal{L}_{se}$ instead of contrastive self-expression loss $\mathcal{L}_{cse}$ in NCAGC ({\bf termed NCAGC w/o CSE}). To be specific, the traditional self-expression loss $\mathcal{L}_{se}$ can be presented by Eq. (\ref{EQ8}). In this scenario, the total loss function of NCAGC w/o CSE can be described by $\mathcal{L}_{rec}+\mathcal{L}_{nbr}+\mathcal{L}_{se}+\mathcal{L}_{coef}$, where $\mathcal{L}_{rec}$ is the loss of node representation reconstruction, $\mathcal{L}_{nbr}$ is the neighborhood contrast loss, $\mathcal{L}_{se}$ is the loss of self-expression, and $\mathcal{L}_{coef}$ is the regularization term. The result of the ablation study on the Contrastive Self-Expression Module is shown in Table \ref{tab:table5}, we can observe that it helps improve the clustering performance on Cora, Citeseer and ACM datasets by 11.3\%, 9.8\% and 25.6\% in terms of ACC, by 10.4\%, 10.6\% and 77.7\% in terms of NMI and by 17.7\%, 21.5\% and 117.8\% in terms of ARI. This result demonstrates that the contrastive self-expression loss $\mathcal{L}_{cse}$ taking the negative sample information into account greatly help to learn a more discriminative self-expression matrix and improve the clustering performance.

Besides, we find that the Contrastive Self-Expression Module could bring more impressive clustering performance compared with Neighborhood Contrast Module, this is because the contrastive self-expression loss $\mathcal{L}_{cse}$ can constrain the learning of the self-expression matrix directly. However, the neighborhood contrast loss $\mathcal{L}_{nbr}$ is used to constrain the learning of node representation and then help to learn a better self-expression matrix indirectly.

{\it 3) Ablation study on attention mechanism:} Since the feature extractor in Symmetric Feature Extraction Module is flexible, we also conduct the ablation study about the effect of the attention mechanism by using GCN instead of GAT for the feature extraction process ({\bf termed NCAGC w/o ATT}). As reported in Table \ref{tab:table5}, we can observe that NCAGC using GAT as the feature extractor surpasses NCAGC w/o ATT on Cora, Citeseer and ACM datasets by 3.6\%, 0.5\% and 1.1\% in terms of ACC, by 7.6\%, 2.6\% and 4.3\% in terms of NMI and by 10.0\%, 1.5\% and 3.1\% in terms of ARI. And it demonstrates that GAT can better encode the node attribute and optimize the node relation with the attention mechanism, which greatly helps to improve the quality of node representation.

\section{Conclusion}
In this paper, we propose an end-to-end framework for attributed graph clustering called NCAGC. To improve the quality of node representation, we design a Neighborhood Contrast Module by maximizing the similarity of neighbor nodes. Besides, to obtain a more discriminative self-expression matrix, we design a Contrastive Self-Expression Module by contrasting the node representation before and after the reconstruction of the self-expression layer. We show that our proposed method achieves state-of-the-art performance comparing with 16 clustering methods over four attributed graph datasets.

\bibliographystyle{IEEEtran}
\bibliography{mybibfile}

\begin{thebibliography}{10}
\providecommand{\url}[1]{#1}
\csname url@samestyle\endcsname
\providecommand{\newblock}{\relax}
\providecommand{\bibinfo}[2]{#2}
\providecommand{\BIBentrySTDinterwordspacing}{\spaceskip=0pt\relax}
\providecommand{\BIBentryALTinterwordstretchfactor}{4}
\providecommand{\BIBentryALTinterwordspacing}{\spaceskip=\fontdimen2\font plus
\BIBentryALTinterwordstretchfactor\fontdimen3\font minus
  \fontdimen4\font\relax}
\providecommand{\BIBforeignlanguage}[2]{{%
\expandafter\ifx\csname l@#1\endcsname\relax
\typeout{** WARNING: IEEEtran.bst: No hyphenation pattern has been}%
\typeout{** loaded for the language `#1'. Using the pattern for}%
\typeout{** the default language instead.}%
\else
\language=\csname l@#1\endcsname
\fi
#2}}
\providecommand{\BIBdecl}{\relax}
\BIBdecl

\bibitem{8579104}
L.~Xu, T.~Bao, L.~Zhu, and Y.~Zhang, ``Trust-based privacy-preserving photo
  sharing in online social networks,'' \emph{IEEE Transactions on Multimedia},
  vol.~21, no.~3, pp. 591--602, 2019.

\bibitem{8613906}
B.~E. Mada, M.~Bagaa, and T.~Taleb, ``Trust-based video management framework
  for social multimedia networks,'' \emph{IEEE Transactions on Multimedia},
  vol.~21, no.~3, pp. 603--616, 2019.

\bibitem{kipf2016semi}
\BIBentryALTinterwordspacing
T.~N. Kipf and M.~Welling, ``Semi-supervised classification with graph
  convolutional networks,'' in \emph{ICLR}, 2017. [Online]. Available:
  \url{https://openreview.net/forum?id=SJU4ayYgl}
\BIBentrySTDinterwordspacing

\bibitem{ying2018graph}
R.~Ying, R.~He, K.~Chen, P.~Eksombatchai, W.~L. Hamilton, and J.~Leskovec,
  ``Graph convolutional neural networks for web-scale recommender systems,'' in
  \emph{ACM SIGKDD}, 2018, pp. 974--983.

\bibitem{9535249}
J.~Yi and Z.~Chen, ``Multi-modal variational graph auto-encoder for
  recommendation systems,'' \emph{IEEE Transactions on Multimedia}, vol.~24,
  pp. 1067--1079, 2022.

\bibitem{cai2018comprehensive}
\BIBentryALTinterwordspacing
H.~Cai, V.~W. Zheng, and K.~C.-C. Chang, ``A comprehensive survey of graph
  embedding: Problems, techniques, and applications,'' \emph{IEEE Transactions
  on Knowledge and Data Engineering}, vol.~30, no.~9, pp. 1616--1637, 2018.
  [Online]. Available: \url{https://doi.org/10.1109/TKDE.2018.2807452}
\BIBentrySTDinterwordspacing

\bibitem{9181470}
Y.~Cai, Z.~Zhang, Z.~Cai, X.~Liu, X.~Jiang, and Q.~Yan, ``Graph convolutional
  subspace clustering: A robust subspace clustering framework for hyperspectral
  image,'' \emph{IEEE Transactions on Geoscience and Remote Sensing}, vol.~59,
  no.~5, pp. 4191--4202, 2021.

\bibitem{DBLP:conf/ijcai/ChengWTXG20}
J.~Cheng, Q.~Wang, Z.~Tao, D.~Xie, and Q.~Gao, ``Multi-view attribute graph
  convolution networks for clustering,'' in \emph{IJCAI}, 2020, pp. 2973--2979.

\bibitem{DBLP:conf/aaai/GaoXWXZ20}
Q.~Gao, W.~Xia, Z.~Wan, D.~Xie, and P.~Zhang, ``Tensor-svd based graph learning
  for multi-view subspace clustering,'' in \emph{AAAI}, 2020, pp. 3930--3937.

\bibitem{hartigan1979algorithm}
J.~A. Hartigan and M.~A. Wong, ``Algorithm as 136: A k-means clustering
  algorithm,'' \emph{Journal of the royal statistical society. series c
  (applied statistics)}, vol.~28, no.~1, pp. 100--108, 1979.

\bibitem{ng2001spectral}
A.~Ng, M.~Jordan, and Y.~Weiss, ``On spectral clustering: Analysis and an
  algorithm,'' in \emph{NeurIPS}, 2001, pp. 849--856.

\bibitem{kipf2016variational}
T.~N. Kipf and M.~Welling, ``Variational graph auto-encoders,'' \emph{arXiv
  preprint arXiv:1611.07308}, 2016.

\bibitem{pan2018adversarially}
\BIBentryALTinterwordspacing
S.~Pan, R.~Hu, G.~Long, J.~Jiang, L.~Yao, and C.~Zhang, ``Adversarially
  regularized graph autoencoder for graph embedding,'' in \emph{IJCAI}, 2018,
  p. 2609––2615. [Online]. Available:
  \url{https://doi.org/10.24963/ijcai.2018/362}
\BIBentrySTDinterwordspacing

\bibitem{2020Learning}
S.~Pan, R.~Hu, S.~F. Fung, G.~Long, J.~Jiang, and C.~Zhang, ``Learning graph
  embedding with adversarial training methods,'' \emph{IEEE Transactions on
  Cybernetics}, vol.~50, no.~6, pp. 2475--2487, 2020.

\bibitem{salehi2019graph}
A.~Salehi and H.~Davulcu, ``Graph attention auto-encoders,'' \emph{arXiv
  preprint arXiv:1905.10715}, 2019.

\bibitem{wang2019attributed}
\BIBentryALTinterwordspacing
C.~Wang, S.~Pan, R.~Hu, G.~Long, J.~Jiang, and C.~Zhang, ``Attributed graph
  clustering: A deep attentional embedding approach,'' in \emph{IJCAI}, 2019,
  pp. 3670--3676. [Online]. Available:
  \url{https://doi.org/10.24963/ijcai.2019/509}
\BIBentrySTDinterwordspacing

\bibitem{bo2020structural}
D.~Bo, X.~Wang, C.~Shi, M.~Zhu, E.~Lu, and P.~Cui, ``Structural deep clustering
  network,'' in \emph{WWW}, 2020, pp. 1400--1410.

\bibitem{wang2021multi}
T.~Wang, J.~Wu, Z.~Zhang, W.~Zhou, G.~Chen, and S.~Liu, ``Multi-scale graph
  attention subspace clustering network,'' \emph{Neurocomputing}, vol. 459, pp.
  302--314, 2021.

\bibitem{9472979}
W.~Xia, Q.~Wang, Q.~Gao, X.~Zhang, and X.~Gao, ``Self-supervised graph
  convolutional network for multi-view clustering,'' \emph{IEEE Transactions on
  Multimedia}, vol.~24, pp. 3182--3192, 2022.

\bibitem{9508843}
Z.~Lin, Z.~Kang, L.~Zhang, and L.~Tian, ``Multi-view attributed graph
  clustering,'' \emph{IEEE Transactions on Knowledge and Data Engineering}, pp.
  1--1, 2021.

\bibitem{DBLP:journals/nn/XiaWYGHG22}
\BIBentryALTinterwordspacing
W.~Xia, S.~Wang, M.~Yang, Q.~Gao, J.~Han, and X.~Gao, ``Multi-view graph
  embedding clustering network: Joint self-supervision and block diagonal
  representation,'' \emph{Neural Networks}, vol. 145, pp. 1--9, 2022. [Online].
  Available: \url{https://doi.org/10.1016/j.neunet.2021.10.006}
\BIBentrySTDinterwordspacing

\bibitem{DBLP:journals/nn/XuXGHG21}
\BIBentryALTinterwordspacing
H.~Xu, W.~Xia, Q.~Gao, J.~Han, and X.~Gao, ``Graph embedding clustering: Graph
  attention auto-encoder with cluster-specificity distribution,'' \emph{Neural
  Networks}, vol. 142, pp. 221--230, 2021. [Online]. Available:
  \url{https://doi.org/10.1016/j.neunet.2021.05.008}
\BIBentrySTDinterwordspacing

\bibitem{9732218}
Y.~Zhang, X.~Zhang, J.~Li, R.~Qiu, H.~Xu, and Q.~Tian, ``Semi-supervised
  contrastive learning with similarity co-calibration,'' \emph{IEEE
  Transactions on Multimedia}, pp. 1--1, 2022.

\bibitem{he2020momentum}
K.~He, H.~Fan, Y.~Wu, S.~Xie, and R.~Girshick, ``Momentum contrast for
  unsupervised visual representation learning,'' in \emph{CVPR}, 2020, pp.
  9729--9738.

\bibitem{chen2020simple}
T.~Chen, S.~Kornblith, M.~Norouzi, and G.~Hinton, ``A simple framework for
  contrastive learning of visual representations,'' in \emph{ICML}, 2020, pp.
  1597--1607.

\bibitem{van2020scan}
W.~Van~Gansbeke, S.~Vandenhende, S.~Georgoulis, M.~Proesmans, and L.~Van~Gool,
  ``Scan: Learning to classify images without labels,'' in \emph{ECCV}, 2020,
  pp. 268--285.

\bibitem{9772930}
T.~Si, F.~He, Z.~Zhang, and Y.~Duan, ``Hybrid contrastive learning for
  unsupervised person re-identification,'' \emph{IEEE Transactions on
  Multimedia}, pp. 1--1, 2022.

\bibitem{9712249}
D.~Cai, S.~Qian, Q.~Fang, J.~Hu, W.~Ding, and C.~Xu, ``Heterogeneous graph
  contrastive learning network for personalized micro-video recommendation,''
  \emph{IEEE Transactions on Multimedia}, pp. 1--1, 2022.

\bibitem{9758652}
Z.~Zhou, Y.~Hu, Y.~Zhang, J.~Chen, and H.~Cai, ``Multiview deep graph infomax
  to achieve unsupervised graph embedding,'' \emph{IEEE Transactions on
  Cybernetics}, pp. 1--11, 2022.

\bibitem{zhu2020deep}
Y.~Zhu, Y.~Xu, F.~Yu, Q.~Liu, S.~Wu, and L.~Wang, ``Deep graph contrastive
  representation learning,'' \emph{arXiv preprint arXiv:2006.04131}, 2020.

\bibitem{zhu2021graph}
------, ``Graph contrastive learning with adaptive augmentation,'' in
  \emph{WWW}, 2021, pp. 2069--2080.

\bibitem{DBLP:conf/nips/YouCSCWS20}
Y.~You, T.~Chen, Y.~Sui, T.~Chen, Z.~Wang, and Y.~Shen, ``Graph contrastive
  learning with augmentations,'' in \emph{NeurIPS}, 2020.

\bibitem{hadsell2006dimensionality}
R.~Hadsell, S.~Chopra, and Y.~LeCun, ``Dimensionality reduction by learning an
  invariant mapping,'' in \emph{CVPR}, 2006, pp. 1735--1742.

\bibitem{you2020graph}
Y.~You, T.~Chen, Y.~Sui, T.~Chen, Z.~Wang, and Y.~Shen, ``Graph contrastive
  learning with augmentations,'' in \emph{NeurIPS}, 2020.

\bibitem{xia2021self}
W.~Xia, Q.~Gao, M.~Yang, and X.~Gao, ``Self-supervised contrastive attributed
  graph clustering,'' \emph{arXiv preprint arXiv:2110.08264}, 2021.

\bibitem{ji2017deep}
P.~Ji, T.~Zhang, H.~Li, M.~Salzmann, and I.~Reid, ``Deep subspace clustering
  networks,'' in \emph{NeurIPS}, 2017, pp. 24--33.

\bibitem{ji2014efficient}
P.~Ji, M.~Salzmann, and H.~Li, ``Efficient dense subspace clustering,'' in
  \emph{IEEE Winter conference on applications of computer vision}, 2014, pp.
  461--468.

\bibitem{mccallum2000automating}
A.~K. McCallum, K.~Nigam, J.~Rennie, and K.~Seymore, ``Automating the
  construction of internet portals with machine learning,'' \emph{Information
  Retrieval}, vol.~3, no.~2, pp. 127--163, 2000.

\bibitem{giles1998citeseer}
C.~L. Giles, K.~D. Bollacker, and S.~Lawrence, ``Citeseer: An automatic
  citation indexing system,'' in \emph{Proceedings of the third ACM conference
  on Digital libraries}, 1998, pp. 89--98.

\bibitem{yang2015network}
C.~Yang, Z.~Liu, D.~Zhao, M.~Sun, and E.~Chang, ``Network representation
  learning with rich text information,'' in \emph{IJCAI}, 2015.

\bibitem{fan2020one2multi}
S.~Fan, X.~Wang, C.~Shi, E.~Lu, K.~Lin, and B.~Wang, ``One2multi graph
  autoencoder for multi-view graph clustering,'' in \emph{WWW}, 2020, pp.
  3070--3076.

\bibitem{wang2017mgae}
C.~Wang, S.~Pan, G.~Long, X.~Zhu, and J.~Jiang, ``Mgae: Marginalized graph
  autoencoder for graph clustering,'' in \emph{ICKM}, 2017, pp. 889--898.

\bibitem{tu2020deep}
W.~Tu, S.~Zhou, X.~Liu, X.~Guo, Z.~Cai, J.~Cheng \emph{et~al.}, ``Deep fusion
  clustering network,'' in \emph{AAAI}, 2021, pp. 9978--9987.

\bibitem{zhang2019attributed}
X.~Zhang, H.~Liu, Q.~Li, and X.-M. Wu, ``Attributed graph clustering via
  adaptive graph convolution,'' in \emph{IJCAI}, 2019, pp. 4327--4333.

\bibitem{park2019symmetric}
J.~Park, M.~Lee, H.~J. Chang, K.~Lee, and J.~Y. Choi, ``Symmetric graph
  convolutional autoencoder for unsupervised graph representation learning,''
  in \emph{CVPR}, 2019, pp. 6519--6528.

\bibitem{velivckovic2017graph}
P.~Veli{\v{c}}kovi{\'c}, G.~Cucurull, A.~Casanova, A.~Romero, P.~Lio, and
  Y.~Bengio, ``Graph attention networks,'' in \emph{ICLR}, 2018.

\end{thebibliography}

\end{document}